\documentclass[10pt,twocolumn,letterpaper]{article}

\usepackage{cvpr}
\usepackage{times}
\usepackage{epsfig}
\usepackage{graphicx}
\usepackage{amsmath}
\usepackage{amssymb}

\usepackage[pagebackref=true,breaklinks=true,letterpaper=true,colorlinks,bookmarks=false]{hyperref}

 \cvprfinalcopy 

\def\cvprPaperID{2897} 

\ifcvprfinal\fi
\begin{document}

\title{End-to-end learning of keypoint detection and matching for relative pose estimation}

\author{Antoine Fond\\
Blippar\\
{\tt\small antoine.fond@blippar.com}
\and
Luca Del Pero\\
Blippar\\
{\tt\small luca.delpero@blippar.com}
\and
Nikola Sivacki\\
Blippar\\
{\tt\small nikola.sivacki@blippar.com}
\and
Marco Paladini\\
Blippar\\
{\tt\small marco.paladini@blippar.com}
}

\maketitle

\begin{abstract}
We propose a new method for estimating the relative pose between two images,
where we jointly learn keypoint detection, description extraction, matching and robust 
pose estimation. While our architecture follows the traditional pipeline for pose estimation 
from geometric computer vision, all steps are learnt in an end-to-end fashion,
including feature matching.  We demonstrate our method for the task of visual localization of a query image within a database of images with known pose. Pairwise pose estimation has many practical applications for robotic mapping, navigation, and AR. For example, the display of persistent AR objects in the scene relies on a precise camera localization to make the digital models appear anchored to the physical environment. We train our pipeline end-to-end specifically for the problem of visual localization. We evaluate our proposed approach on localization accuracy, robustness and runtime speed. Our method achieves state of the art localization accuracy on the 7 Scenes dataset.
\end{abstract}

\section{Introduction}
Our work fits in the domain of visual localisation, i.e. the problem of estimating in real-time the camera pose of a query image within a known map of the environment. We focus on visual localization methods that use an RGB database of images with associated pose, and a single RGB image as query. In this paper, we present a novel method to jointly solve for keypoints detection, keypoints matching and camera pose estimation between a database reference image and a live query image. We assume that the environment has been mapped using either Simultaneous Localization and Mapping (SLAM) or Structure from Motion (SFM) which recovers the 6 DoF camera pose for each RGB image and the associated 3D structure in a consistent 3D coordinate system.

Traditional solutions to the problem of relative pose estimation rely on four steps: detecting keypoints, computing their descriptors, a data association step to match keypoints across the two images, and a robust pose estimation step to compute the pose that minimizes the reprojection of the matched points, while discarding outliers. Localization methods typically use hand-crafted features \cite{place-reco-survey2016, Sattler2011ICCV, li2012worldwide}. Recent work has proposed replacing some of those steps with learning methods \cite{dymczyk2018landmarkboost, Zagoruyko_2015_CVPR, MatchNet2015} or avoid computing keypoints altogether \cite{Kendall_2015_ICCV, Zhan_2018_CVPR, shotton2013}. We propose a novel architecture that still explicitly performs all the traditional four steps, but learns them all jointly in an end-to-end fashion. Our contributions are as follows:
\begin{itemize}
\item End-to-end learning of all the traditional four steps of relative pose estimation
\item A novel trainable layer for feature matching
\item A self-supervised data augmentation method to generate ground-truth matches from RGB-D images
\end{itemize}
We evaluate our method on the indoor localization benchmark 7 Scenes \cite{glocker-ismar-2013} which comprises of sequences of registered RGB-D images for training and for testing in a variety of environments with a variety of camera trajectories.

\begin{figure*}[!h]
\centering
\includegraphics[width=0.8\textwidth]{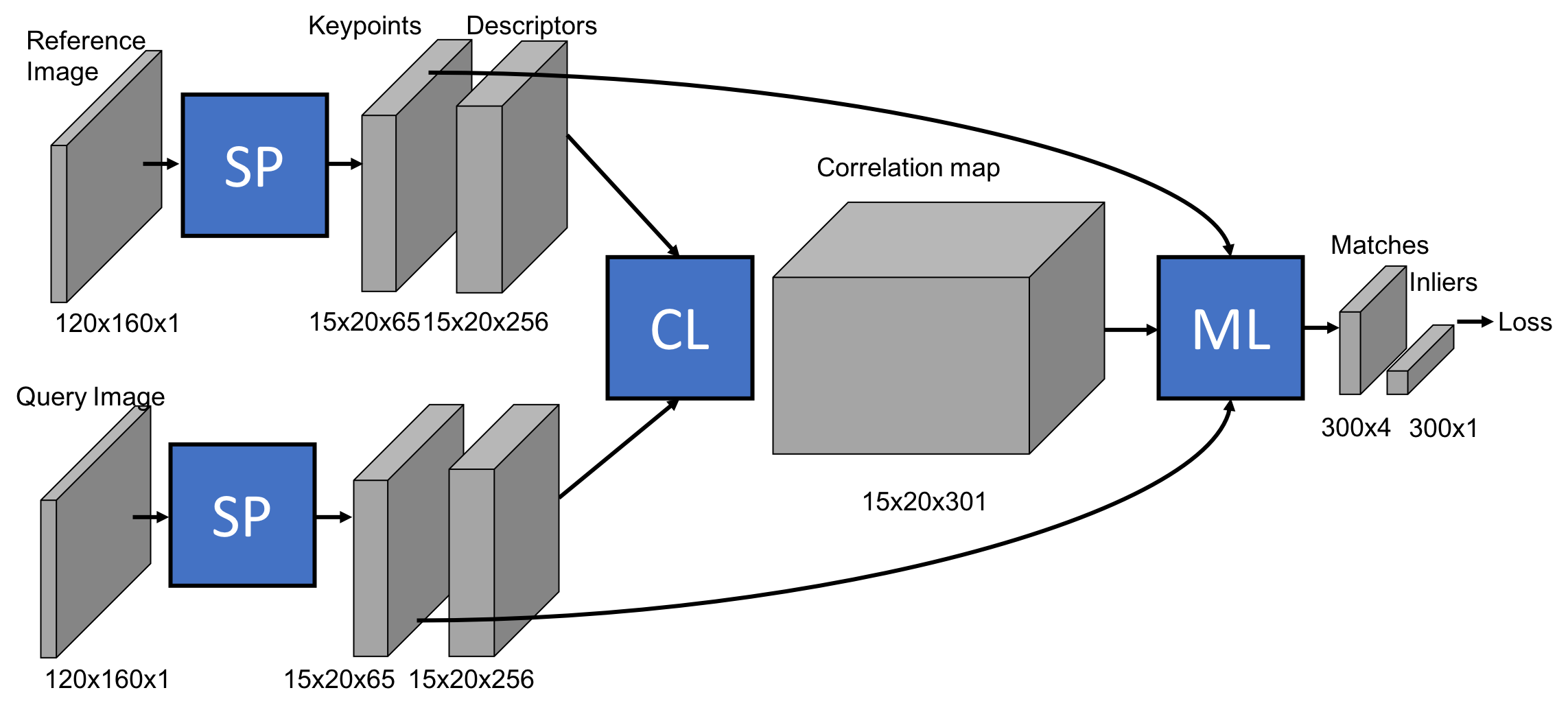}
\caption{\label{fig:fig_overiew} Overview of the model, images are single-channel grayscale, the superpoint \cite{superpoint} layers (SP) outputs keypoints and descriptor tensors, the keypoint indices are propagated to the matching layer (ML) trough softargmax, while the descriptors are used in the cross-correlation layer (CL). The 4D cross-correlation tensor is mapped into a 3D tensor via pseudo-hilbert curve mapping (see Figure \ref{fig:fig_hilbert})}
\end{figure*}

\section{Related work}
Visual localization is a fundamental problem for robotics, autonomous vehicles and augmented reality where the robot or device needs to rely on the current view from the camera to compute its pose. The goal is to obtain the camera pose relative to a known map of the environment. Visual localization methods typically rely on distinctive local features or whole-image descriptors \cite{place-reco-survey2016}. Sattler \etal \cite{Sattler2011ICCV} demonstrated the use of 2D-3D matching and robust pose estimation in large 3D maps. Li \etal \cite{li2012worldwide} use a planet-wide point cloud with image feature descriptors for place recognition and precise localization.

Visual localization methods that do not rely on large 3D maps typically employ image retrieval techniques such as NetVLAD \cite{arandjelovic2016netvlad}. Such retrieval methods compute whole-image descriptors, which allows matching a query image to a large database of geo-tagged images, without using a 3D model \cite{sattlercvpr17}. Camposeco \etal \cite{Camposeco2018CVPR} proposed to leverage both 2D-2D feature matching between images and 2D-3D matches followed by robust pose estimation.

Machine learning methods have been employed to solve the keypoint detection and matching problems. For example, Han \etal \cite{MatchNet2015} proposed a learning method to jointly solve feature extraction and matching, Mishchuk \etal \cite{DBLP:journals/corr/MishchukMRM17} have used deep learning to compute feature descriptors, and DeTone \etal \cite{superpoint} proposed a network architecture to extract keypoints and descriptors that could run on devices with low power consumption. Brachmann \etal \cite{brachmann2017dsac} solve the robust pose estimation problem via machine learning via differentiable RANSAC (DSAC). Alternative localization methods that employ end-to-end learning without explicit feature extraction learn to predict pose directly such as PoseNet \cite{Kendall_2015_ICCV}. Shotton \etal \cite{shotton2013} propose a localization method using RGB-D images that does not rely on keypoint matching, but directly regress the pose of a query image within a global model of the environment.

Our approach is learning based, but it preserves the traditional, \textbf{interpretable} steps of keypoint extraction and matching. We learn this end-to-end by maximising inlier count and minimising the pose error using a set of training RGB-D images with known pose. We show that on the problem of visual localization our method achieves better speed and robustness than traditional methods.

\section{Proposed method}

\begin{figure}[!h]
\centering
\includegraphics[width=\linewidth]{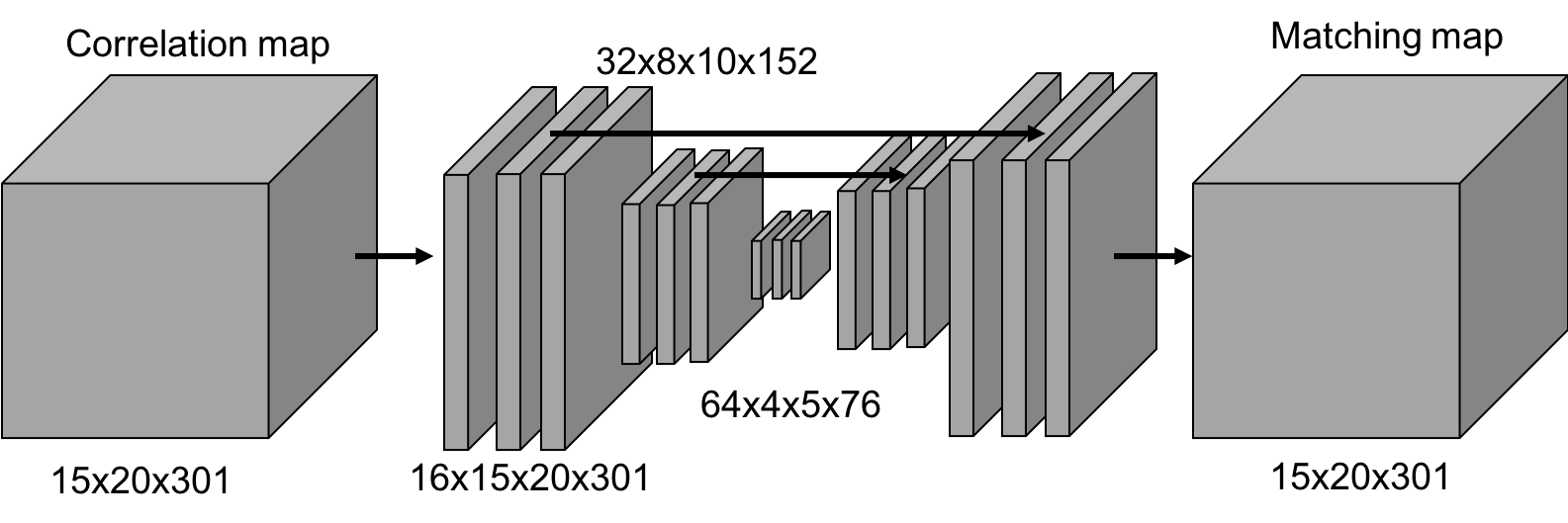}
\caption{\label{fig_network}Details of the V-Net \cite{milletari2016v} encoder-decoder process  used inside the matching layer to compute the matching	 map}
\end{figure}

\subsection{Architecture}

Our novel architecture recovers the pose of a query image $I_q$ given another reference image $I_r$ of the same scene from which the pose and geometry is known. This typically requires 4 steps: keypoints extraction, local descriptors computation, matching and robust pose estimation. We propose a novel CNN architecture for end-to-end relative pose estimation that explicitly performs the traditional four steps above.  In this section we will show how we integrate the first three steps into one single differentiable network. Fig. \ref{fig:fig_overiew} pictures an overview of the network. Both images are first processed through Superpoint \cite{superpoint} to obtain keypoints and descriptors, then the 2 descriptor tensors are fused using a correlation layer \cite{dosovitskiy2015flownet}. From the output of this correlation layer our novel matching layer produces points correspondences that are used to predict the pose from which the final loss function is computed using \cite{dang2018eigendecomposition}. Last, we discuss how to use our network for localization. (Sec. \ref{sec:pose})

\subsubsection{Superpoint}

For the keypoints and descriptors extraction we use the encoder-decoder architecture from Superpoint \cite{superpoint}. In Superpoint, the image is subdivided in a regular grid of cells (patches) each $8 \times 8$ pixel, and the network extracts two semi-dense feature maps. The first submap $K$ encodes the position of the keypoints, the second submap $D$ their descriptors. Hence, for a grayscale input image of size $\left( H,W \right)$, the outputs of both decoders are respectively $\left( \frac{H}{8} \times \frac{W}{8} \times 65 \right)$ for the keypoints $K\left( i,j \right)$ and $\left( \frac{H}{8} \times \frac{W}{8} \times 256 \right)$ for the descriptor $D\left( i,j \right)$. The $65=8 \times 8 + 1$ channel dimensions specifies the keypoint position (or its absence) within each patch. Softmax is applied channel-wise ignoring the 65th channel in the normalization for the keypoint map and the descriptor map is normalized using L2. Moreover each patch is also associated with one single descriptor of size $256$.

\subsubsection{Correlation layer}

Given the query and the reference image we first compute both keypoint and descriptor maps respectively $\left( K_q,D_q \right)$ and $\left( K_r,D_r \right)$ using Superpoint \cite{superpoint}. Following the way matching is done classically we compute the dot product between all pairs of descriptors from the 2 images. This operation is differentiable and sometimes referred to as correlation layer \cite{dosovitskiy2015flownet}. A high correlation score between descriptors can be interpreted as a putative match. In our case this score is also weighted by the keypoint score $K' \left( i,j \right) = 1-K \left( i,j,65 \right)$ of both descriptors leading to the following 4D tensor:

\begin{equation}
C_{q,r} \left( i,j,i',j' \right) = K_q' \left( i,j \right) K_r' \left( i',j' \right) D_q \left( i,j \right)^T D_r \left( i',j' \right) 
\end{equation}

\subsubsection{Matching layer}

While we could follow the standard matching process by associating each patch $\left( i,j \right)$ with its highest correlated patch $\left( i',j' \right)$ in $C_{q,r}$, we aim to learn the structure of that 4D tensor so to find better matches.
Our intuition is that a set of matches is good if they are spatially consistent, for example two putative matches are more likely to be reliable if their points land close to each other in both query and reference image.
While this idea of exploiting the spatial relationship of putative matches has been formalised in previous work through graph theory \cite{leordeanu2005spectral,torresani2008feature}, we aim to learn the local structure in that 4D space using a CNN.
Thus the matching layer produces a sparse map by convolving the 4D correlation map with a set of filters, which we train to generate peaks at positions $\left( i,j,i',j' \right)$ if $\left( i,j \right)$ and $\left( i',j' \right)$ should be matched together. This has the potential to leverage global information, while traditional
descriptor matching is inherently local.

\begin{figure}[!h]
\centering
\includegraphics[width=0.8\linewidth]{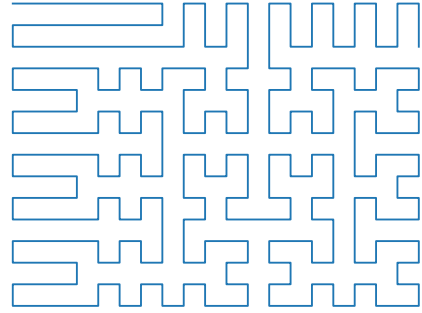}
\caption{Pseudo-Hilbert's curve $\mathcal{H}$ for $H=15,W=20$}
\label{fig:fig_hilbert}
\end{figure}

\begin{figure}[!h]
\centering
\includegraphics[width=\linewidth]{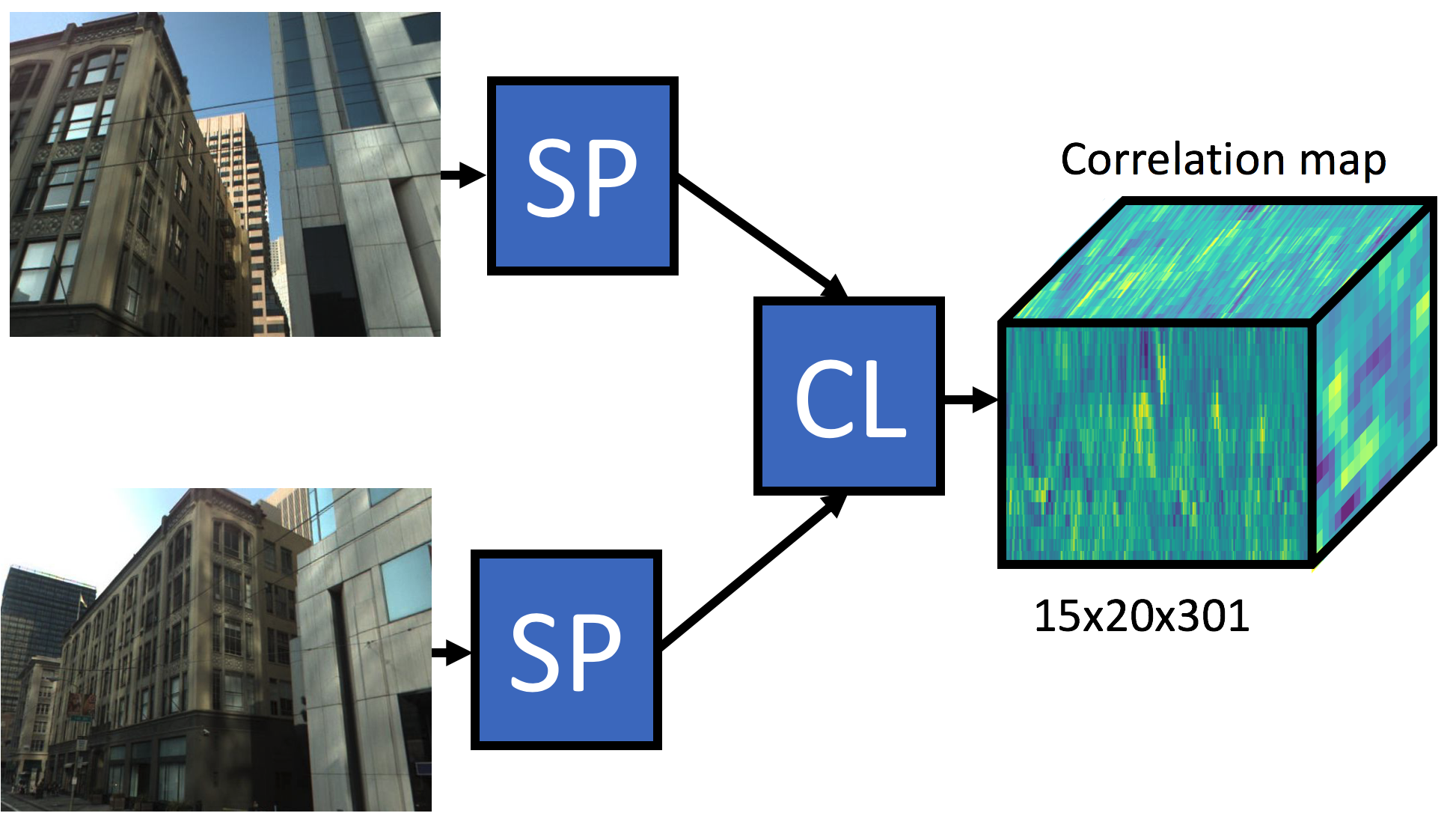}
\centering
\includegraphics[width=\linewidth]{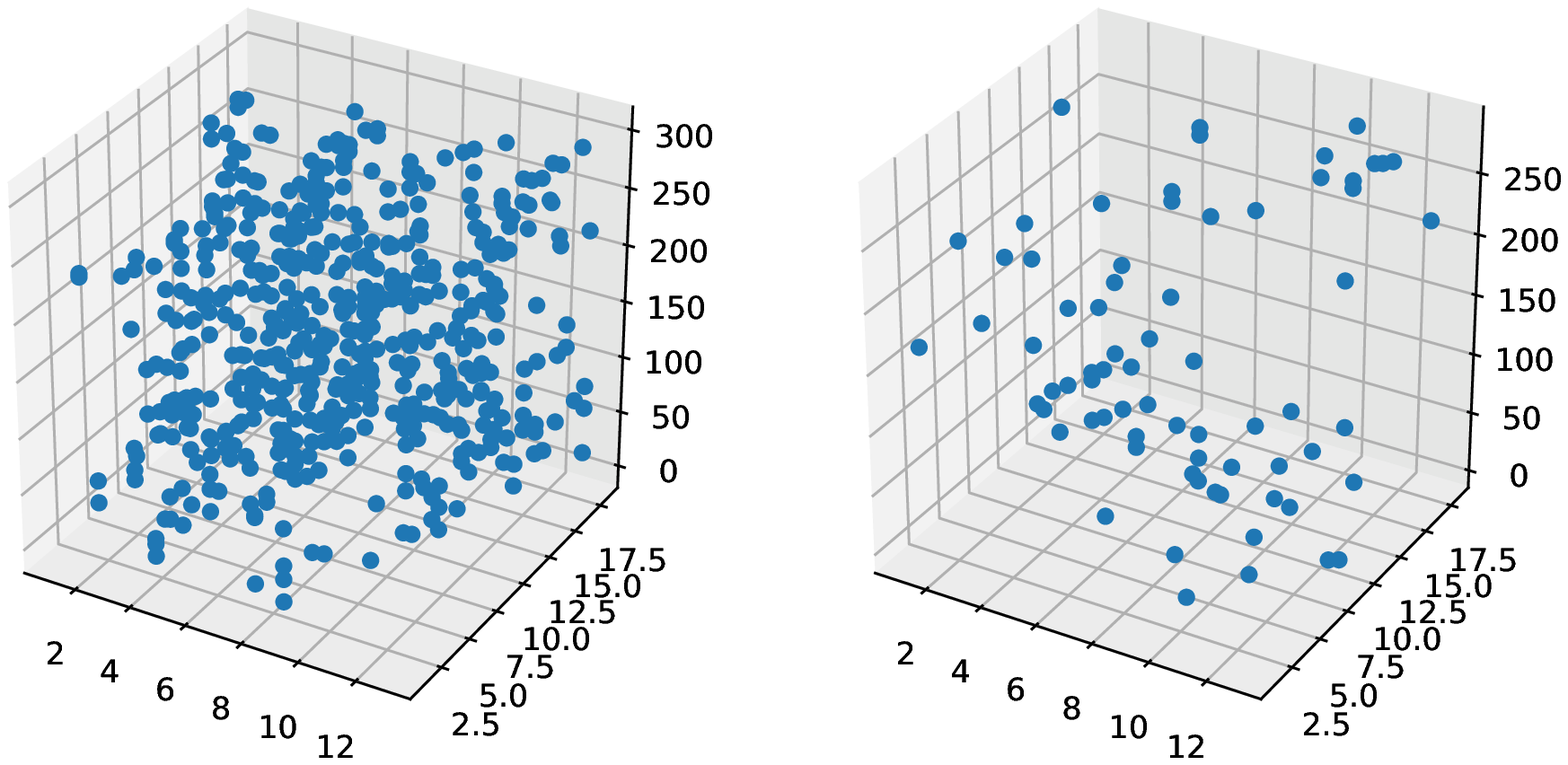}
\caption{Example of local maxima in the 3D correlation map, First row: Computing the correlation map using Superpoint (SP) and the correlation layer (CL). Bottom row: local maxima in the correlation map. Left: line-by-line raveling, right: Hilbert's curve raveling shown in Figure \ref{fig:fig_hilbert}}
\label{fig:fig_hilbert_spatial}
\end{figure}

In practise and for efficiency the 4D tensor from the correlation layer is ravelled into a 3D tensor. Thus the indices $\left( i',j' \right)$ are transformed into a single index $k$ following a 2D Hilbert's curve $\mathcal{H}$ \cite{zhang2006pseudo} (Fig. \ref{fig:fig_hilbert}) to better preserve spatial locality (Fig. \ref{fig:fig_hilbert_spatial}). Instead of expensive 4D convolutions we can then apply 3D convolutions on that 3D tensor of size $\left( H,W,H \times W \right)$. Actually some keypoints in image $I_q$ may have no matching keypoint in image $I_r$. This situation is handled with an extra dustbin class (or "no-match" class) in the matching layer output $M$. The dimensions of that matching map $M$ are then $\left( H,W,H \times W +1 \right)$. Processing 3D convolutions does not change the number of channels so we add an extra zeros-filled dimension to the input correlation map to be of the same size $\left( H,W,H \times W +1 \right)$.

To capture the local structure at different scale we use a similar encoder-decoder architecture as in Vnet \cite{milletari2016v} which is a generalisation of Unet \cite{ronneberger2015u} to 3D tensors. The connections between the encoder and decoder stages of the same resolution enable finegrained details in the matching map which is important for the overall accuracy. The networks details can be seen in figure Fig. \ref{fig_network}.

At this stage peaks in 3D tensors represent keypoints and matches. To be able to train the network end-to-end so that an actual pose can be predicted from those keypoints and matches we need an additional step to convert the peaks into correspondences between points coordinates $\left( kp\left( i,j \right), kp' \left( i,j \right) \right)$. This is done here using softargmax \cite{chapelle2010gradient} and through the following steps:

\begin{equation}
\begin{split}
kp \left( i,j \right)_x = 8 \times i + \sum_{m,n} m s_K \left( m,n,i,j \right) \\
kp \left( i,j \right)_y = 8 \times j + \sum_{m,n} n s_K \left( m,n,i,j \right) \\
\text{with } s_K \left( m,n,i,j \right) = K \left( i,j,8 \times m+n \right)
\end{split}
\label{eq:coordinates}
\end{equation}


\begin{equation}
\begin{split}
kp' \left( i,j \right)_x = \sum_{i',j'} s_M \left( i,j,i'j' \right) kp \left( i',j' \right)_x \\
kp' \left( i,j \right)_y = \sum_{i',j'} s_M \left( i,j,i'j' \right) kp \left( i',j' \right)_y \\
\text{with } s_M \left( i,j,i'j' \right) = \frac{\exp \left( M \left(i,j,\mathcal{H} \left( i',j' \right) \right) \right)}{\sum_{n=1}^{H \times W} \exp \left( M \left( i,j,n \right) \right)}
\end{split}
\label{eq:coordinates_matches}
\end{equation}

\subsection{Losses}

The last part of a keypoint-based method for localization generally aims to find the pose from the set of correspondences. As some correspondences can be outliers this step is usually done using a robust estimator such as RANSAC \cite{fischler1981random}. In \cite{dang2018eigendecomposition}, the authors proposed a practical alternative to DSAC \cite{brachmann2017dsac} to make that loss differentiable. The loss is derived from the solution of the Weighted Direct Linear Transform and made of 3 components: $\tilde{e}$ that depends on the ground-truth relative pose $\left( R_{r,q},T_{r,q} \right)$, $X$ that depends on the correspondences $q_{i,j} = \left( kp\left( i,j \right), kp_n' \left( i,j \right) \right)$ and a weight for each correspondence. The formula are described in the section about the PnP problem from \cite{dang2018eigendecomposition} using the same notations as here. We adapted that loss to our problem by defining the correspondences weights as $M' \left( i,j \right) = 1-M \left( i,j, H \times W + 1 \right)$ (i.e. the inverse of the dustbin class) and by dropping the regularization term.

\begin{equation}
L_{\text{pose}} \left( M' \right) = \tilde{e}^T X^T M' X \tilde{e}
\end{equation}

To avoid the trivial solution where all weights drop to zero, we use instead a differentiable version of the number of inliers using a sigmoid function $\sigma$. Indeed as we assumed the depth and global pose of both images to be known, we can compute the reprojection of the keypoints from the reference image $kp' \left( i,j \right)$ into the query image. Let $C_{\text{calib}}$ denotes the camera calibration matrix and $P_{r,q} = \left( R_{r,q} | T_{r,q} \right)$ the relative pose between the images then the reprojection can be written $\mathcal{R} \left( kp' \left( i,j \right) \right) = C_{\text{calib}} P_{r,q} kp_n' = C_{\text{calib}} P_{r,q} \frac{C_{\text{calib}}^{-1} kp' \left( i,j \right)}{\left\|\ C_{\text{calib}}^{-1} kp' \left( i,j \right) \right\|} \text{depth}_r \left( i,j \right)$.

\begin{equation}
\begin{split}
& L_{\text{inliers}} \left( W \right) = \exp \left( - \kappa s_i \right) \\
& \text{with } s_i = \sum_{i,j} M'\left( i,j \right) \sigma \left( \tau -\left\|\ kp \left( i,j \right) - \mathcal{R} \left( kp' \left( i,j \right) \right) \right\|  \right)
\end{split}
\end{equation}

Eventually, we noticed that without any additional constraint keypoints tend to converge to a dense solution where there is a keypoint in each patch centered in the middle of the patch even if it means losing some accuracy in pose. To overcome that we added a unitary term that enforce keypoints to be close to the keypoint position detected by Superpoint \cite{superpoint}. That additional loss $L_{\text{keypoints}}$ is the sum of the cross-entropy loss on the keypoints map $K_q$ and $K_r$ with Superpoint's \cite{superpoint} keypoints being the ground-truth.

The final loss is a weighted sum of those losses with hyperparameters $\alpha$, $\beta$. Empirically we have noticed that $\alpha=\beta=2$ give good results. We also use $\kappa=0.1$ and $\tau=16$ for the other hyperparameters.

\begin{equation}
L_{\text{final}} =  L_{\text{pose}} + \alpha L_{\text{inliers}} + \beta L_{\text{keypoints}}
\end{equation}

\subsection{Training}

\subsubsection{Pairs generation}

To train our end-to-end approach we need to generate pairs $\left( I_q, I_r \right)$ with known relative pose and depth. We use the Stanford 2D-3D Semantic dataset \cite{armeni2017joint} that provides 70496 RGB images with corresponding depth and global pose. 46575 were used for training (area 2,4 and 5) and the rest for testing. We propose here a procedure to find pairs of them (with their relative pose) that are suitable for training our method. Note that, since this dataset contains only global poses, we cannot just use all possible pairs in the dataset (e.g. we cannot use a pair where the images are in different rooms, or have no overlap in the field of view). Here we outline a procedure to find pairs suitable for training.

For each training image $I_t$ we then use the Pitts250k pre-trained NetVLAD \cite{arandjelovic2016netvlad} to retrieve the 64 closest images $I_r$. As the global pose is known for each image, we first discard the poses among those 64 retrievals that are incompatible with the current training image. Two poses are incompatible if their relative distance is more than 20m and if there is no intersection between the 2 fields of view. The use of a retrieval step beforehand instead of simply checking the incompatibility between all pairs is twofold. First it deals with occlusions as some parts may not be visible (because of walls or furnitures) even if their pose are compatible. Second, it biases the training to be closer to the localization use-case where we first get the references images from image retrievals using NetVLAD \cite{arandjelovic2016netvlad}. 


Eventually we remove images pairs which have very close poses $\left( \left\|\ T_{t,r} \right\|\ \leq 0.5m \text{ and } \angle R_{t,r} \leq 5^{\circ} \right)$ to force the network to solve non trivial cases for pose estimation (i.e. wide baseline).

\subsubsection{Scene adaptation}

Instead of using Superpoint's \cite{superpoint} keypoints as ground-truth for the keypoint unitary loss $L_{\text{keypoints}}$, we propose to extend the concept of Homography Adaptation introduced in Superpoint \cite{superpoint} to produce more repeatable keypoints over different viewpoints.
We call that self-supervision scheme Scene adaptation and it simply consists of aggregating points from reprojections of different viewpoints of the scene using ground-truth poses and depth rather than through random homography.

In practise we use the same generated training pairs for the different viewpoints. However it can happen that a keypoint visible in one image is not visible in the other because of occlusion. We discard those situations by comparing the depth of the reprojected keypoint $\text{depth}_q \left( C_{\text{calib}} P_{r,q} kp_n' \right)$ with its z-value before reprojection $\left( 0,0,1\right) . P_{r,q} kp_n'$. 


Following the original idea of Homography Adaptation, these new keypoints are iteratively updated during training. It means that every 20 epochs, we recompute the keypoints ground-truth used in $L_{\text{keypoints}}$ by aggregating across viewpoints the keypoints extracted using the last learned state of the network.

\subsubsection{Implementation details}

We use Pytorch framework with ADAM solver for learning. We run the optimization over $100$ epochs with a learning rate of $10^{-5}$ and mini-batches of size $32$. To overcome the non-linearity of the softmax layers and sigmoid function, we first train the network with a learning rate multiplier of $0$ on the keypoints encode-decoder part initialized with Superpoints's \cite{superpoint} weights. These weights being fixed the network is forced to focus on learning the filters of the matching layer. Then in a second phase, the network is trained end-to-end (i.e. including the keypoints) with the weights of the matching layer initialized to those of that first phase.

\subsection{\label{sec:pose}Pose refinement}

To use our network on the problem of localising a query image in a database of images with known pose, as in most retrieval methods based on keypoints, we first find the closest $N=16$ putative database images to the query using NetVLAD \cite{arandjelovic2016netvlad}. We then predict the pose between the query and each putative image using our network, and keep the pose $P_{\text{max}}$ with most inliers.
 
At this stage, we perform a further refinement step, where we add correspondences from the putative images whose predicted pose is not too far from $P_{\text{max}}$ (i.e. $\left\|\ T_{r} - T_{\text{max}} \right\|\ \leq 1m$). Last, we recompute the pose on this larger set of matches using P3P-RANSAC and refine it by non-linear optimization on the inliers using Levenberg-Marquardt. This step improves the accuracy of our method, and has a low computational cost because our network extracts few keypoints with a high inlier ratio on which RANSAC is very efficient.

\section{Experiments}

\subsection{\label{sec:stanford}Results on relative pose estimation}

We first evaluate our architecture on relative pose estimation on a test set from Stanford 2D-3D Semantic dataset \cite{armeni2017joint}. We generate 50000 image pairs from the 23921 images of areas 1,3,6. In this experiment we aim to show that our matching layer improves matching compared to the standard matching process of keypoint-based methods. For this, we use standard keypoint matching as a baseline, i.e. we match keypoints in the first image to keypoints in the second image with minimum descriptor distance and under Lowe's ratio constrains. Note that also the baseline uses Superpoint \cite{superpoint} for keypoints and descriptors for fair comparison. We also compare to Superpoint without its post-process step. Superpoint  \cite{superpoint}  uses a post-processing step that, unravels the keypoint map to the size of the input image and applies Non Maximum Suppression (NMS) to extract the keypoints after the forward pass. The descriptors are also interpolated at those locations from the neighbouring patches.  We compare to Superpoint with and without the post-processing step. We note that post-processing takes 20\% of the running-time and, since it is not differentiable, it cannot benefit from end-to-end learning.

\begin{figure*}[!h]
\centering
\includegraphics[width=0.485\linewidth]{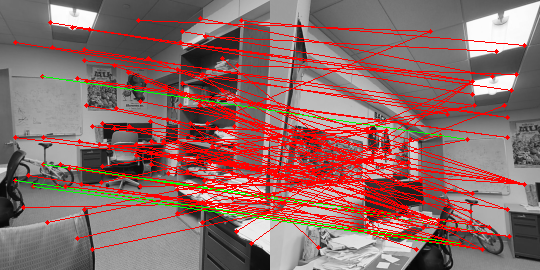} \hspace{0.001\linewidth} \includegraphics[width=0.485\linewidth]{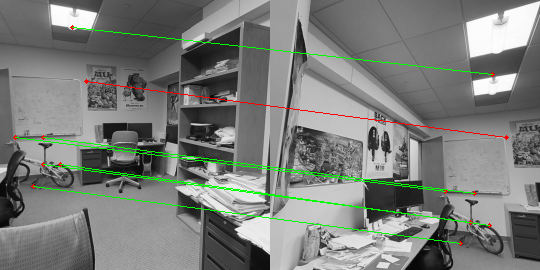} \\
\includegraphics[width=0.485\linewidth]{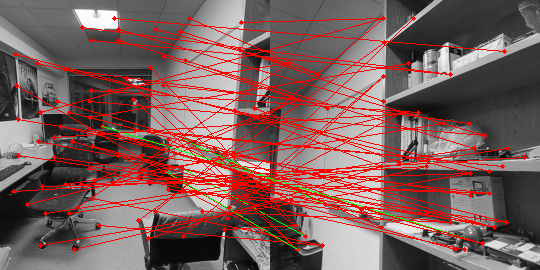} \hspace{0.001\linewidth} \includegraphics[width=0.485\linewidth]{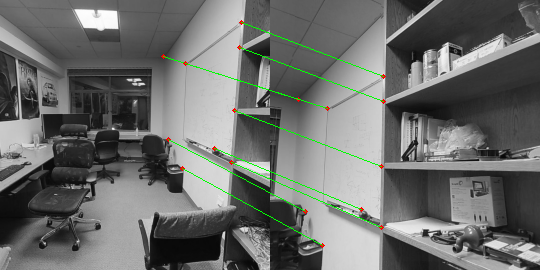}
\caption{Examples of keypoint matching (Sec \ref{sec:stanford}) with Superpoints (left) and our method (right). The inliers from the best pose estimated by RANSAC are in green and the outliers in red.}
\label{fig:fig_inliers}
\end{figure*}

We choose two different metrics to evaluate the improvement of our matching layer. The first is the inliers-ratio $\alpha$. An inlier is a keypoint $kp$ whose the corresponding keypoint $kp'$ in the other image is reprojected sufficiently close to $kp$ ($\epsilon \leq 8$ pixels) using the ground-truth relative pose and depth. The inliers-ratio is the ratio between the number of inliers and the total number of correspondences. Table \ref{tab:tab_stan} shows that our method predict much more reliable matches with an average inliers-ratio of $\overline{\alpha}=0.76$ which is more than 2 times higher than Superpoints \cite{superpoint} with standard matching (Fig. \ref{fig:fig_inliers}). That allow us to make only about 16 RANSAC iterations instead of about 400 with the same confidence of finding the inliers.

The second metric is the error between the ground truth relative pose and the pose estimated from the correspondences. The pose is estimated using P3P-RANSAC and refined on inliers using non-linear optimization. $N_{\%}$ corresponds to the percentage of poses that have been estimated by P3P-RANSAC (i.e. pairs with at least 4 correspondences). Table \ref{tab:tab_stan} shows here that our method is more robust as it always gives at least 4 matches while Superpoints fails in 4\% of the cases. $r_a$ and $r_m$ are respectively the average and median error of the rotation in degree. Likewise $t_a$ and $t_m$ are respectively the average and median error of the translation in meters. With few number of inliers and a small inliers-ratio, RANSAC may fail to estimate the best pose, all samples being equivalent. However the better matches of our methods enables a more accurate estimation in average. Also thanks to the better localized keypoints, our method also outperforms the baseline in accuracy when RANSAC succeeds which is measured with the gain in median error. That gain in accuracy is even bigger if we consider Superpoint \cite{superpoint} without its post-process.

\begin{table}[h!]
\centering
\begin{tabular}{|c|c|c|c|}
  \hline
  & ours & SP \cite{superpoint} w NMS & SP \cite{superpoint} wo NMS \\
  \hline
  $\overline{\alpha}$ & $0.76$ & $0.37$ & $0.36$ \\
   $\mathbf{N_{\%}}$ & $100$ & $96$ & $94$ \\
  $\mathbf{r_a}$ & $5.06^{\circ}$ & $11.66^{\circ}$ & $24.54^{\circ}$ \\
  $\mathbf{r_m}$ & $0.73^{\circ}$ & $0.76^{\circ}$ & $1.23^{\circ}$ \\
  $\mathbf{t_a}$ & $1.8$ & $49$ & $\infty$ \\
  $\mathbf{t_m}$ & $0.06$ & $0.07$ & $0.12$ \\
  \hline
\end{tabular}
\caption{\label{tab:tab_stan}Results on Stanford 2D-3D pairs (Sec. \ref{sec:stanford}). $\overline{\alpha}$: average inlier ratio, $N_{\%}$: percentage of queries estimated (i.e.  pairs with at least 4 correspondences) $r_a$, $r_m$: average and median rotation error respectively (degrees). $t_a$, $t_m$: average and median translation error in meters. Superpoint performance improve with a Non Maximum Suppression (NMS) Our method shows better accuracy as well as robustness.}
\end{table}

\subsection{Results on localisation}

We also evaluate our method using our efficient localization pipeline (Sec. \ref{sec:pose}) on the 7scenes dataset \cite{glocker-ismar-2013} which is a standard benchmark for indoor localization. The dataset consists of images with associated pose and depth from several tracks. It is split into a training set that we use for as references and a test set we use for queries.

The pipeline takes query image $I_q$ as input and first retrieve the $N$ closest references $\left\{ I_{r,j} \right\}_{1 \leq i \leq N}$. Knowing the depth and global pose of those retrieval references, we compute the global pose of the query using P3P-RANSAC using the keypoint correspondences predicted by our method. Thus we select either the query global pose from the one corresponding to the maximum number of inliers among the retrieval or with pose refinement (PR) to measure the impact of that last step. We constrains the pipeline to make it practicable for a real-time application. Particularly we use a small number of retrieval $N=16$ and very low-resolution images $\left( 160 \times 120 \right)$.

\begin{table*}[h!]
\small
\centering
\begin{tabular}{|c|c|c|c|c|c|c|c|c|}
  \hline
  & NV+ours+PR & NV+ours & NV+SP\cite{superpoint}  & NV+ORB\cite{rublee2011orb} & NV+SIFT\cite{Lowe2004} & AS \cite{sattler2017efficient} & PoseNet\cite{Kendall_2015_ICCV} & DSAC\cite{brachmann2017dsac} \\
  \hline
  Chess & $3$/$\mathbf{5}$ & $4$/$7$ & $5$/$7$ & $4$/$12$ & $3$/$\mathbf{5}$ & $4$/$-$ & $13$/$-$ & $\mathbf{2}$/$-$ \\
  Fire & $4$/$\mathbf{8}$ & $5$/$12$ & $6$/$11$ & $6$/$13$ & $5$/$9$ & $\mathbf{3}$/$-$ & $27$/$-$ & $4$/$-$ \\
  Heads & $\mathbf{2}$/$\mathbf{4}$ & $3$/$5$ & $3$/$5$ & $6$/$\infty$ & $5$/$\infty$ & $\mathbf{2}$/$-$ & $17$/$-$ & $3$/$-$ \\
  Office & $\mathbf{4}$/$\mathbf{9}$ & $5$/$\mathbf{9}$ & $6$/$\mathbf{9}$ & $7$/$\infty$ & $5$/$\infty$ & $9$/$-$ & $19$/$-$ & $\mathbf{4}$/$-$ \\
  Pumpkin & $7$/$\mathbf{76}$ & $9$/$154$ & $11$/$200$ & $11$/$\infty$ & $10$/$\infty$ & $8$/$-$ & $26$/$-$ & $\mathbf{5}$/$-$ \\
  Red. kit. & $\mathbf{4}$/$\mathbf{11}$ & $5$/$16$ & $6$/$20$ & $7$/$\infty$ & $6$/$\infty$ & $7$/$-$ & $23$/$-$ & $5$/$-$ \\
  Stairs & $13$/$\mathbf{114}$ & $16$/$224$ & $16$/$\infty$ & $21$/$\infty$ & $15$/$\infty$ & $\mathbf{3}$/$-$ & $35$/$-$ & $117$/$-$ \\
  \hline
\end{tabular}
\caption{\label{tab:tab_seven_translation}Median and average translation errors on the 7Scenes \cite{glocker-ismar-2013}. The values are respectively median error in translation (cm) / average error in translation (cm). NV stands for NetVLAD \cite{arandjelovic2016netvlad}, PR is our pose refinement step, SP stands for Superpoint \cite{superpoint} and AS for ActiveSearch \cite{sattler2017efficient}}
\end{table*}

\begin{table*}[h!]
\small
\centering
\begin{tabular}{|c|c|c|c|c|c|c|c|c|}
  \hline
  & NV+ours+PR & NV+ours & NV+SP\cite{superpoint}  & NV+ORB\cite{rublee2011orb} & NV+SIFT\cite{Lowe2004} & AS \cite{sattler2017efficient} & PoseNet\cite{Kendall_2015_ICCV} & DSAC\cite{brachmann2017dsac} \\
  \hline
  Chess & $1.3$/$1.9$ & $1.8$/$2.8$ & $1.9$/$2.7$ & $1.6$/$5.1$ & $\mathbf{1.2}$/$\mathbf{1.8}$ & $1.9$/$-$ & $4.5$/$-$ & $\mathbf{1.2}$/$-$ \\
  Fire & $1.8$/$\mathbf{3.9}$ & $2.3$/$4$ & $2.5$/$\mathbf{3.9}$ & $2.4$/$9.5$ & $2.1$/$5.1$ & $\mathbf{1.5}$/$-$ & $11.3$/$-$ & $\mathbf{1.5}$/$-$ \\
  Heads & $1.7$/$3.8$ & $2.1$/$3.7$ & $2.4$/$3.6$ & $2.4$/$9.5$ & $\mathbf{1.4}$/$\mathbf{2.8}$ & $\mathbf{1.4}$/$-$ & $13.0$/$-$ & $2.7$/$-$ \\
  Office & $1.5$/$\mathbf{2.2}$ & $1.8$/$2.9$ & $1.8$/$2.6$ & $2.2$/$9.7$ & $\mathbf{1.4}$/$2.8$ & $3.6$/$-$ & $5.5$/$-$ & $1.6$/$-$ \\
  Pumpkin & $2.2$/$\mathbf{5.4}$ & $2.8$/$6.7$ & $2.7$/$8.7$ & $2.7$/$21.6$ & $2.4$/$15.7$ & $3.1$/$-$ & $4.5$/$-$ & $\mathbf{2.0}$/$-$ \\
  Red. kit. & $\mathbf{1.5}$/$\mathbf{3.8}$ & $1.8$/$5.3$ & $1.9$/$8.4$ & $2.2$/$13.3$ & $1.8$/$5.5$ & $3.3$/$-$ & $5.3$/$-$ & $2.0$/$-$ \\
  Stairs & $3.5$/$\mathbf{8.4}$ & $3.9$/$12$ & $4.4$/$11.7$ & $5.1$/$28.2$ & $4.3$/$14.2$ & $\mathbf{2.2}$/$-$ & $12.4$/$-$ & $33.1$/$-$ \\
  \hline
\end{tabular}
\caption{\label{tab:tab_seven_angles}Median and average rotation errors on the 7Scenes \cite{glocker-ismar-2013}. The values are respectively median error in rotation (degrees) / average error in rotation (degrees). NV stands for NetVLAD \cite{arandjelovic2016netvlad}, PR is our pose refinement step, SP stands for Superpoint \cite{superpoint} and AS for ActiveSearch \cite{sattler2017efficient}}
\end{table*}

For fair comparisons we focused on methods that shows real-time compatibility. We specifically compare to other keypoint-based approaches to estimate the pose within the same pipeline. That includes Superpoint with standard matching, ORB \cite{rublee2011orb} and SIFT \cite{Lowe2004} with 
Lowe's criteria on matches. SIFT is also used for direct 2D-3D matching with ActiveSearch \cite{sattler2017efficient}. For approaches that extracts SIFT or ORB features we use the full resolution image $\left( 640 \times 480 \right)$ as with low-resolution there were to many failure cases. We also compare to deep-learning methods that are fast but requires a specific training per scene when in our case the same network is used for all the 7 scenes.

The poses from Stanford 2D-3D dataset \cite{armeni2017joint} are very biased. Because they comes from sparsely located panorama there is much less variation in translation than in rotations. To compensate for that lack of data, we finetune our network on the concatenated training data of 7Scenes \cite{glocker-ismar-2013}. Our method shows competitive results on 7 scenes \cite{glocker-ismar-2013} (Tables \ref{tab:tab_seven_translation} and \ref{tab:tab_seven_angles}). We especially outperforms all other keypoint-based methods even with low-resolution input both in accuracy (median error) and robustness (average error).

The efficiency of the method is measured by its average running time of 167 ms per query. This is possible by the use of low-resolution images which is also important for an online localization service as it reduces data transfer. We also make use of the ability of the framework to process the $16$ retrieved images in parallel by loading them into one batch.

\section{Conclusions}

We have proposed an end-to-end learning method that explicitly includes the 4 standards part of a keypoint-based localization pipeline: keypoints extraction, description, matching and pose estimation. Our method achieves state-of-the-art results on standard benchmark for localization among keypoint-based methods, significantly outperforming other end-to-end learning approaches. Its efficiency makes it suitable for real-time applications, for example Augmented Reality. We note that it can be very powerful when combined with odometry methods like Visual Inertial Odometry.

Our method requires poses and depth for training. While we evaluated on RGB-D data, note that our method could be used also on RGB datasets after applying dense SLAM or SfM methods. In future work, we will investigate whether out network can be trained in an unsupervised fashion from data without explicit pose and depth ground truth.
{\small
\bibliographystyle{ieee}
\bibliography{localisation}
}

\end{document}